# Explainable Failure Prediction and Prevention in Maritime

**Dionisis Kalogeropoulos[1], Georgia Sovatzidi[1], Panagiotis G. Kalozoumis[1], Dimitris K. Iakovidis[1]**

[1]*Dept. of Computer Science and Biomedical Informatics,*
*University of Thessaly*
{dkalogerop, gsovatzidi, pkalozoumis, diakovidis}@uth.gr

**Abstract**

Maritime systems operate in highly dynamic environments where unexpected equipment failures can compromise safety, reliability, and operational efficiency. Recent advances in artificial intelligence (AI), machine learning, digital twins, and predictive maintenance enable proactive failure prediction and prevention. However, ensuring trustworthy and explainable decision-making remains a major challenge in safety-critical maritime applications. This chapter reviews key AI technologies required for explainable failure prediction and prevention in maritime systems and presents a conceptual architecture capable of supporting autonomous or human-in-the-loop corrective actions. This architecture integrates data acquisition, time-series forecasting, anomaly detection, risk assessment, decision-making, and explainable AI into a closed-loop framework. With reference to the architectural components, a review and discussion of relevant maritime studies is performed, outlining their methods, advantages, and limitations. Furthermore, it highlights current challenges, including uncertainty and robustness, model generalization, explainability, limited availability of maritime datasets, and operational deployment, and identifies future research directions toward trustworthy AI-assisted maritime decision-making.

# 1. Introduction

The maritime industry enables the movement of over 80% of world trade by volume, underlining its critical role in many sectors, including the global economy. The marine machinery and complex systems included in modern ships operate under demanding and often unpredictable environmental conditions. To ensure their optimal performance and safe operation, proper design and regular maintenance of the corresponding equipment are required. However, unexpected equipment failures often occur, which can be associated with serious consequences, *e.g.*, ship malfunctions, financial losses, safety hazards, as well as environmental problems [1]. A recent analysis of maritime accidents that occurred from 2014 to 2023 revealed that the dominant cause of failures was mainly due to technical issues and, in particular, machinery damage or malfunctions, highlighting the need to develop and implement maintenance strategies [2]. These strategies include the following: a) Reactive Maintenance (RM), which addresses equipment failures after they occur, leading to unpredictable and often costly downtime; b) Preventive Maintenance (PM), which is performed on predetermined schedules, frequently resulting in unnecessary repairs and overscheduled interventions, thereby introducing substantial operational expenses regardless of the actual condition of the equipment [1]; c) Predictive Maintenance (PdM), which leverages data and advanced analytics to forecast when a component is likely to fail, enabling maintenance activities to be performed proactively before failure occurrence. PdM can significantly minimize unplanned downtime, reduce maintenance costs by 20-50%, and effectively extend the operational lifespan of critical equipment [3].

The integration of Artificial Intelligence and Machine Learning (ML) has profoundly revolutionized PdM, enhancing the performance, accuracy, autonomy, and adaptability of PdM systems within complex and dynamic operational environments. This transition aligns well with the "Shipping 4.0" paradigm, which envisions the transformation of conventional vessels into "smart ships" through the integration of the Internet of Ships (IoS) and AI. It promises automated fleet management, further reductions in downtime, and improved compliance with strict environmental regulations [4]. The evolution from RM to PdM, powered by AI, represents a fundamental shift from merely "fixing things when they break" to "proactively preventing breakage". AI-driven PdM is essential for optimal vessel operation. As the maritime industry transitions towards greater autonomy, ensuring that onboard systems operate within desired parameters and maximizing equipment availability becomes of paramount importance [2].

Several studies have investigated the use of AI for PdM. Specifically, Kalafatelis *et al*. [4] detailed the latest data-driven PdM techniques applied across various vessel systems, including propulsion, electrical, auxiliary, and hull components. They found that ML-driven methods are highly suitable for PdM due to their ability to learn from complex datasets, adapt without detailed physical models, make real-time predictions, and be robust to noisy data. They also noted that Deep Learning (DL) models offer superior performance but require extensive labeled datasets, making federated learning and transfer learning promising alternatives. Toh and Park [5] summarized DL methods for vibration-based Structural Health Monitoring (SHM) and fault monitoring. They categorized studies based on vibration factors and provided an interpretation of Deep Neural Networks (DNNs). They highlighted that DL methods are increasingly applied due to their flexibility and robustness, achieving high accuracy, using automatic feature extraction from data, and are well-suited for complex nonlinear systems. A survey in 2024 [6] reviewed AI applications for product quality control and PdM in the era of Industry 4.0. They discussed the use of AI to anticipate machine failures, optimize maintenance schedules, and estimate the Remaining Useful Life (RUL) of the machinery. Data imbalance, lack of explainability, and domain dependency were identified as key limitations. Ucar *et al*. [1] reviewed recent developments in AI-based PdM. They focused on key components (sensors, data preprocessing, AI algorithms, decision-making modules), trustworthiness, and future trends. They found that AI enhances the performance, accuracy, autonomy, and adaptability of PdM systems, significantly reducing unplanned downtime and extending equipment lifespan. A review by Durlik *et al*. [7] highlighted the potential of AI to transform maritime safety. They reviewed AI applications in maritime safety and risk management, including case studies of Maersk, DNV GL (Veracity platform), and Wärtsilä (SmartPredict system). They reported that AI systems can identify potential hazards, provide real-time decision support, monitor hazardous materials, predict equipment failures, and optimize shipping routes, enhancing overall safety and operational efficiency. Garcia *et al*. [8] discussed data-driven PdM programs. They found that combining high-frequency sensing with ML analytics can cut downtime by 10-20%, leading to global savings of 8-15 billion USD per year while also reducing energy waste and CO2 emissions. Lei *et al*. [9] systematically reviewed current mainstream industrial fault diagnosis methods from the perspective of industrial big data. They found that undetected faults increase maintenance costs and downtime, and that data-driven methods are crucial for industrial digital intelligence transformation. Furthermore, they identified the increasing difficulty in acquiring diverse and complete data as a significant challenge.

Shipping accidents and their potential consequences point to the shared responsibility of various stakeholders, including shipping companies and crew members. Over the past decades, advancements in ship design, technology, regulatory frameworks, and risk management systems have led to a 70% reduction in reported shipping accidents [10]. However, while the frequency of marine accidents and risks may be decreasing, a single incident can have devastating and long-term consequences for marine ecosystems [11]. A systematic investigation into the causes of maritime hazards and accidents was performed using the Domino effect model, which offered a systematic approach to understanding the causes of accidents [12]; it envisioned accidents as a sequence of events, represented by five key elements: a) social environment, b) human error (negligence), c) unsafe acts or mechanical and physical conditions, d) accidents, and e) injuries. Mitigating or addressing any of these five key elements can contribute to the prevention of accidents in maritime operations.

Although research on marine hazards has focused mainly on naval technology, human error has been recognized as the main cause of marine accidents and several methods have been developed to address them [13][14]. Such errors are due to human weaknesses and, as crews become increasingly dependent on new technologies, situational awareness problems are increasing significantly and have a negative impact on many sectors, including the aquatic environment [15]. For example, it was estimated that 30–50% of oil spills were caused directly or indirectly by human error [16]. A detailed review of studies using ML in maritime risk analysis in terms of context, datasets, and methodologies has been presented in [17]. Several studies analyzed accidents in various sea areas, *e.g.,* Yangtze River [18], Taiwan [19], and China [20], and conducted visibility estimation for safer navigation in the strait of Istanbul [21], Mississippi River [22], and arctic waters [23]. In [15], a statistical analysis of historical data and the assessment of the safety level of all basic merchant ship types were presented, in terms of accident occurrence, initial frequencies, and basic consequences. For each ship type, several accident cases were also presented with respect to accident category, total losses of ships, and number of fatalities. In [13], 50 years of research publications were summarized and analyzed from different aspects, revealing the evolution of maritime accident research. A literature review of over 180 papers published until 2018 was presented in [24], presenting different threats and their consequences related to the maritime industry as well as various risk assessment (RA) models and computational algorithms.

This chapter presents a novel, generic, and modular conceptual architecture for explainable failure prediction and prevention in maritime, designed to be applicable to various monitored components (mechanical or electronic) and communication channels. The proposed design provides a flexible framework for identifying potential deterioration states of a vessel component by leveraging their respective sensor data streams and other functional information channels and business constraints, such as operational, maintenance, ship lifecycle/historical performance, weather, and environmental data, increasing operational efficiency while reducing the possibility of a hazard. Furthermore, the methodology combines multiple forecasting and detection ML models with robust decision making and RA pipelines to deliver valuable actuation healing strategies ("heal" in the sense of recovering from faults and regaining normative performance) and suggestions that mitigate potential future failures. Overall, the modular design makes the architecture easily adaptable to different types of ships consisting of various types of components and its design is inherently reliable as it provides explainable outcomes that can be understood by humans and gain their trust.

The remainder of this chapter is organized as follows: Section 2 provides a background on relevant AI concepts; Section 3 introduces the conceptual architecture for explainable failure prevention; Section 4 presents data acquisition and processing systems; Section 5 provides a literature review on methods utilized for time-series forecasting and anomaly detection; Section 6 analyzes recently proposed approaches for decision-making in maritime; Section 7 identifies research challenges from this review study, and finally, Section 8 summarizes the derived conclusions.

# 2. Background

Considering the undesirable consequences of maritime accidents, highlighted in Section 1, being able to predict and prevent failures in maritime is of great importance. AI can substantially contribute to this direction by providing methodological tools for: a) time-series forecasting, b) anomaly detection, c) decision-making, and d) explainability.

## 2.1 Time-Series Forecasting

Predictive models in maritime operations focus on forecasting critical parameters and potential failure points before they materialize into serious incidents [6]. These models leverage historical operational data, sensor readings, environmental conditions, and maintenance records to predict equipment degradation, system failures, and hazardous situations. By analyzing patterns in engine performance, hull stress measurements, navigation data, and weather forecasts, predictive algorithms can estimate the RUL of components, anticipate mechanical breakdowns, and forecast dangerous operating conditions. This proactive approach enables vessel operators and maintenance crews to schedule interventions during optimal windows, preventing unexpected failures at sea.

Time-series forecasting modeling involves analyzing historical time-ordered data with computer algorithms that leverage statistical learning theory to create automated solutions that uncover hidden temporal patterns and behaviors in data. These methodologies are generally categorized by their mathematical framework as probabilistic or non-probabilistic, and by their functional form, as linear, nonlinear, or ensemble/hybrid. Non-probabilistic models focus on point estimation, where the goal is to minimize a specific loss function (*e.g.,* mean squared error). Within this domain, linear approaches remain foundational. Linear regression is a representative approach that models the relationship between a dependent variable and one or more independent variables by fitting a linear equation to observed data. Penalized linear models, such as Ridge and Lasso, extend standard regression by adding a regularization term to the loss function. Ridge regression uses $L_2$ regularization (sum of squares of coefficients) to handle multicollinearity, while Lasso uses $L_1$ regularization (absolute value of the sum of coefficients) to perform feature selection by shrinking some coefficients to zero. Another standard but effective approach is Kalman filtering, which estimates the state of a dynamic system from a series of noisy measurements. It is widely used for real-time tracking and signal processing [25].

Probabilistic models provide a full probability distribution for future values, allowing for uncertainty quantification through prediction intervals. Examples include autoregressive methods, such as AutoRegressive Integrated Moving Average (ARIMA), which handles non-stationary data by assessing the changes between consecutive values, Vector Autoregression (VAR) captures the linear interdependencies among multiple time series, and Gaussian Processes (GP), which is a non-parametric Bayesian approach that defines a distribution over functions. GPs are highly flexible and provide inherent uncertainty estimates. Bayesian linear regression [26] constitutes another representative probabilistic model that treats the parameters as probability distributions rather than fixed points, incorporating prior knowledge into the estimation process. Furthermore, probabilistic Kalman filters [27] extend the standard Kalman filter by explicitly modeling the evolution of the entire probability density of the system state.

While linear models leverage multiple affine transformations to map historical observations to future values, they often struggle with complex, real-world dynamics. Nonlinearity can be introduced into these models through feature engineering, where raw inputs are preprocessed and transformed into nonlinear feature spaces. A more natural approach to the non-linearity aspect comes in the form of artificial neural networks (ANNs). Specifically, ANNs mimic the neural network of the human brain by modeling the neurons and their activations as parameters and activation functions, respectively. The most common ANN is the multilayer feedforward NN, which consists of consecutive layers of neurons propagating the information from their input to the output, while transforming and mapping the data to vector spaces of different dimensions. The adaptation of the parameters of an ANN with multiple (>3) neural layers to a training dataset, is known as DL. ANN architectures that can naturally capture temporal relations include the recurrent neural networks and their variants, such as the gated recurrent unit (GRU) or long short-term memory (LSTM) networks [28]. In these networks, a hidden state (memory) enables them to remember the order that the inputs appear, while capturing nonlinear relations to improve forecasting capabilities. Furthermore, advances in the domain include Transformer architectures that can provide a different communication mechanism between data points (feature vectors) via the attention concept. Their computational requirements are high; however, their effectiveness has been demonstrated in a variety of relevant applications [29], [30] . Other baseline approaches that can effectively capture linear and nonlinear relations, such as support vector machines (SVR) with their regression variants, regression trees (Random Forests, boosting) and ensembles have been proven effective and efficient [31].

Subsequent advances in DL have introduced more advanced architectures. One notable example is the Generative Adversarial Network (GAN), which is based on the idea of adversarial training. A GAN consists of two components: a generator, which creates synthetic data sequences, and a discriminator, which tries to distinguish between real and generated data. Through this competition, the generator learns to produce more realistic data. In forecasting applications, this process can improve data quality and diversity, helping models better generalize and handle variability in temporal data [32]. While GANs focus on improving data representation through generation, another complementary direction is to learn how to make sequential decisions directly from data. This is the focus of Reinforcement Learning (RL), where an agent interacts with an environment and learns to optimize its behavior over time by maximizing a cumulative reward. In forecasting settings, RL can be used to define proxy tasks that encourage the discovery of temporal dependencies between data points and long-term patterns [33].

The performance of ML-based forecasting models may vary considerably depending on the scope and availability of data. Data preprocessing, feature engineering, and the selected model can be affected by the level of available prior knowledge. In this context, prior knowledge is typically provided by domain experts-individuals with an understanding of the underlying system or process (*e.g.,* engineers, analysts, or medical doctors). Based on the level of such knowledge, ML settings can be categorized into four types [34]: i) no prior information about anomalies is available; ii) only normal behavior is known; iii) labeled examples of known anomalies are available; and iv) anomalies can be generated based on predefined rules. Based on the above categorization, the following four categories of methods are distinguished: (I) unsupervised, (II) semi-supervised, (III) supervised and (IV) self-supervised methods, respectively. In unsupervised learning, no labeled (annotated) data are available, and the model relies solely on the structure of the data to identify patterns and detect deviations that may correspond to anomalies. In semi-supervised learning, only examples of normal behavior are provided, allowing the model to learn a notion of normality and flag deviations from it as potential anomalies, although in some settings a small number of labeled anomalies may also be available. In supervised learning, both normal and anomalous instances are labeled, enabling the model to directly learn how to distinguish between them, often leading to strong performance when sufficient data exists. Finally, in self-supervised learning, labels are not explicitly given but are derived from the data through the design of proxy tasks, such as predicting missing or future values, with anomalies identified when the model fails to perform these tasks accurately. By combining these concepts, a system can be created with effective ways of forecasting future situations, which can signal an unwanted or unexpected future behavior that could deviate from the notion of operational normalcy, hindering or even stopping maritime operations.

## 2.2 Anomaly Detection

Anomaly detection (AD) is the process of identifying data points, events, or patterns that deviate significantly from the expected behavior of a system or the norm of a dataset. Expected behavior is commonly described using terms with positive connotations, such as normal, usual, or regular, and terms with negative connotations, such as abnormal, unusual, irregular, rare, deviations, or outliers are used to describe observations that differ from this behavior [35]. In general, anomalies refer to patterns in data that do not conform to a well-defined notion of normal behavior. Examples of anomalies include the degradation of an engine component in PdM and unresponsive redundant systems in fault and hazard prevention, where there is a significant deviation from normal system operation, increasing the risk of operational instability.

AD frameworks are essential for identifying critical events across data-intensive domains, where the vast majority of data is considered normal. Real-world AD problems can be diverse, and there are several types of anomalies that have been identified in the literature [36], [37], making the problem non-trivial and hard to generalize. These types can be categorized into broader categories, including: i) point anomalies, ii) conditional or contextual anomalies [38], iii) group or collective anomalies [39], iv) logical anomalies, which can be considered as an extended version of (i)-(iii) [40], v) low-level sensory anomalies, and vi) high-level semantic anomalies [41] which are largely studied in the context of DL methods. Typically, AD models try to classify data points into two classes: "normal" and "anomaly", making it a form of binary classification. Thus, the most direct way to train an AD model is through binary classification with empirical risk minimization, using normal data and data with anomalies as the two classes [31].

Early approaches to anomaly detection were based primarily on statistical tests; these approaches typically construct a mathematical model representing normal data behavior and then apply inference testing to determine whether new instances deviate from that baseline. Several methods are used to conduct statistical AD, such as proximity-based, parametric, non-parametric, and semi-parametric methods. Recently, an increasing number of works employ ML algorithms to automate the process of knowledge acquisition from examples by constructing models that distinguish between ordinary and abnormal classes. In recent years, ML have been widely applied to many AD tasks by integrating NNs which can learn complex and hierarchical latent features expressing intricate data distributions [42]. These ML-based AD approaches rely on different learning strategies to identify deviations from normal behavior. Reconstruction-based methods [43] learn to reproduce normal data patterns and flag anomalies when reconstruction errors are high, while template-based methods [44] compare new observations against predefined or learned reference patterns to detect mismatches. Furthermore, synthesis-based methods [45] generate synthetic data to augment the available dataset, increasing its size and diversity and helping the model better distinguish between normal and anomalous cases. Additionally, contrastive learning [46] has been used to learn representations by bringing similar data points closer and pushing dissimilar ones apart, improving the separation between normal and anomalous patterns. Knowledge distillation [47] transfers knowledge from a larger or more complex model to a simpler one, enabling efficient AD without significant loss in performance. Building on these advances, recent progress in computer science has led to a paradigm shift with the emergence of foundation models (FMs) [48], which leverage large-scale pretraining to learn general representations that can be adapted to a wide range of ML tasks, including AD.

Recent advances in computer science have created a paradigm shift driven by the emergence of FMs. These models exhibit impressive few-shot or even zero-shot generalization capabilities across a broad spectrum of downstream tasks, such as AD or time-series forecasting, often surpassing task-specific models. In a few-shot setting, the model adapts to a new task using only a small number of labeled examples, while in a zero-shot setting, it can generalize to new tasks without seeing any labeled examples at all, relying instead on the knowledge acquired during large-scale pretraining. Within this evolving landscape, there are two main categories of works: (i) adapting large language models (LLMs) for the task of AD, and (ii) leveraging FMs pre-trained on large-scale specific domain data to support a variety of applications, such as directly prompting an LLM to identify anomalies in a dataset by a textual prompt. These algorithms, when applied to real-world data and situations, can determine the probability of a potential anomalous incident that might arise to disrupt a maritime operation. This, combined with the vast increasing interconnectivity of systems-of-systems, gives rise to a multitude of AD that provide indicators of the operational state of a vessel and, naturally, are in need of aggregation by a decision-making mechanism.

## 2.3 Decision Making and Risk Assessment

In the context of maritime decision-making RA refers to the process of evaluating multiple risk factors, including static risk factors, *e.g*., the ship type, as well as dynamic ones, *e.g.,* whether, to determine the level of risk associated with a vessel or operation. Then, this information is used to guide actions like inspections, routing, or safety interventions [49]. RA is a systematic process of collecting, analyzing, and managing data to determine the probability and potential impacts of adverse events in a given context. In the maritime domain, RA involves the evaluation of various operational, environmental, and human factors to assign an appropriate level of risk to specific scenarios, such as navigational hazards, equipment misfunctions, and/or marine accidents. The RA process is typically structured in three main stages: risk identification, analysis, and assessment. During risk identification, potential risk factors are systematically identified and are quantified to assess the severity of these factors, often by incorporating historical data or expert knowledge. Risk analysis is a systematic process used to identify, evaluate and describe risks for the purpose of decision-making. In the context of maritime, risk analysis may involve assessing the probability and consequences of accidental marine incidents, such as collisions or groundings, to improve safety and guide management decisions.

This stage often incorporates historical data, statistical models, simulation techniques, and expert knowledge to assess the severity and uncertainty of the risk [50]. Furthermore, by applying qualitative and quantitative approaches, such as models based on fuzzy logic and/or subjective probabilities, the uncertainty and imprecision, which are often inherent in expert knowledge, can be limited. Fuzzy logic enables the representation of uncertainty and vagueness in complex systems and facilitates human-like reasoning through linguistic variables [51]. Representative examples of fuzzy models are Fuzzy Cognitive Maps, whose graphical structure provides transparency and interpretability, based on their cause-and-effect relationships and using linguistic terms, allowing users to understand the entire reasoning process, until reaching a final prediction [52]. During the risk assessment, the identified risk factors are compared and evaluated based on predetermined criteria and/or thresholds, aiming to determine their acceptability and apply relevant actuations, if needed, to mitigate the risk.

## 2.4 Explainable Artificial Intelligence

Despite their effectiveness in solving complex problems in various contexts, the majority of ML approaches are considered as "*black boxes",* as their internal procedures are not understood by humans. As a result, user trust in these systems as well as their use in more critical areas, including maritime operations, is limited. Considering the need to adopt and understand ML models, XAI has seen significant growth in recent years [52]. Specifically, XAI focuses on providing comprehensible explanations of *how* a model works and *why* it makes certain predictions [53]. Recent research has addressed the interpretability of black-box ML models in the maritime domain, aiming to provide transparent decision-making frameworks for predictive safety and organizational workflows [11]. For the majority of maritime industries, the explainability of AI models is essential for building trust in a system. Considering PdM, it is important for engineers and managers to understand why an ML model recommends specific actions, such as part replacement or repair of a vessel component; incorrect or unjustified maintenance decisions may lead to unnecessary downtime, increased operational costs, or critical system failures.

Furthermore, explainability constitutes a key factor in European regulatory approaches [53]; the General Data Protection Regulation (GDPR) and the European (EU) AI Act emphasize the importance of XAI, requiring the explainability of the "black box" models that are developed, thus allowing users and stakeholders to understand the rationale of the decision-making process. Therefore, XAI can contribute to the transparency of decision-making processes, allowing stakeholders to assess fairness, verify compliance with legal standards and determine the consequences of algorithmic decisions [54].

# 3 A Conceptual Architecture for Explainable Failure Prevention

To address the need for explainable failure prevention in maritime, this section presents a generic approach applicable to the prediction and prevention of failures in monitored maritime mechanical or electronic components and communication channels. Knowledge and expertise from stakeholders (*e.g*., ship operators and crew) can be incorporated, *e.g*., as rules, into knowledge-based decision-making systems, to automatically infer useful information for failure prevention. Such information may include the monitoring of one or more composite sea vessel components, *i.e.*, assemblies made up of multiple interconnected parts working together, and possible actions that need to be taken to prevent the failure. The failure prevention process is explainable, providing users with comprehensible results, thereby contributing to enhancing their trust in the system.

A ship includes multiple components, each with different properties and characteristics, such as the main engine, motors, pumps, and steering gear. These components are part of the ship's various systems, including communication, navigation, electrical, and mechanical systems, which are accessible to the ship crew for operation and maintenance. The risk of system or component failure at different parts of a ship may vary; for instance, a malfunction in the main engine of a vessel may evoke a significantly higher risk than a failure in a pump. Therefore, monitoring and data acquisition (DAQ) via sensors is necessary to effectively prevent failure.

A schematic representation of the conceptual architecture for explainable failure prevention is presented in Fig. 1(a). The proposed architecture is fully modular, composed of multiple interconnected nodes, set up for local monitoring, failure prediction, and decision making in different parts of a sea vessel, as illustrated in Fig. 1(b). Let us consider a monitored component *C* associated with a specific part of a ship system of interest, which is monitored by a set of sensors (Fig. 1(a)). First, DAQ is performed using one or more sensors and utilizing data that may come from different channels, *e.g.,* data from the ship's communication and information systems; data processing is considered as part of the DAQ block. The acquired data are streamed to an AD system that analyzes the data aiming to identify the existence of anomalies. In parallel, a time-series prediction system forecasts future signal values, whose outputs are used to support the AD system. Specifically, the AD system recognizes anomalies and extracts semantic information about the status of the ship component and future failures that may occur, either from the current state or from forecasted values. If an anomaly is detected in a ship component, the corresponding information is provided as input into a decision-making algorithm designed to infer the failure risk in the ship and propose respective control instructions to limit and/or prevent it. The implementation of inferred actions depends on the confidence in the inferred decisions. Decisions are only implemented if confidence is high, which indicates low uncertainty based on the available data; otherwise, the process is repeated until a reliable result is obtained. Uncertainty in the decision-making process may arise from multiple sources, including the uncertainty propagated through streamed data, the AD system, and the time-series forecasting system. Furthermore, the decision-making module is also able to infer control instructions for failure risk limitation and prevention with higher confidence. The proposed architecture also includes an explanation system that can provide explanations of the decisions derived from this model, in a way understandable to users, aiming to gain their trust. The explanations can be provided in various forms, including *e.g.,* fuzzy IF-THEN rules that can explain the relations existing among the data, using cause-and-effect relations. Based on the suggested instructions, the actuator, which can be a system or a human from the supervisory personnel, can take healing measures to prevent failure. The entire procedure follows a closed-loop approach.

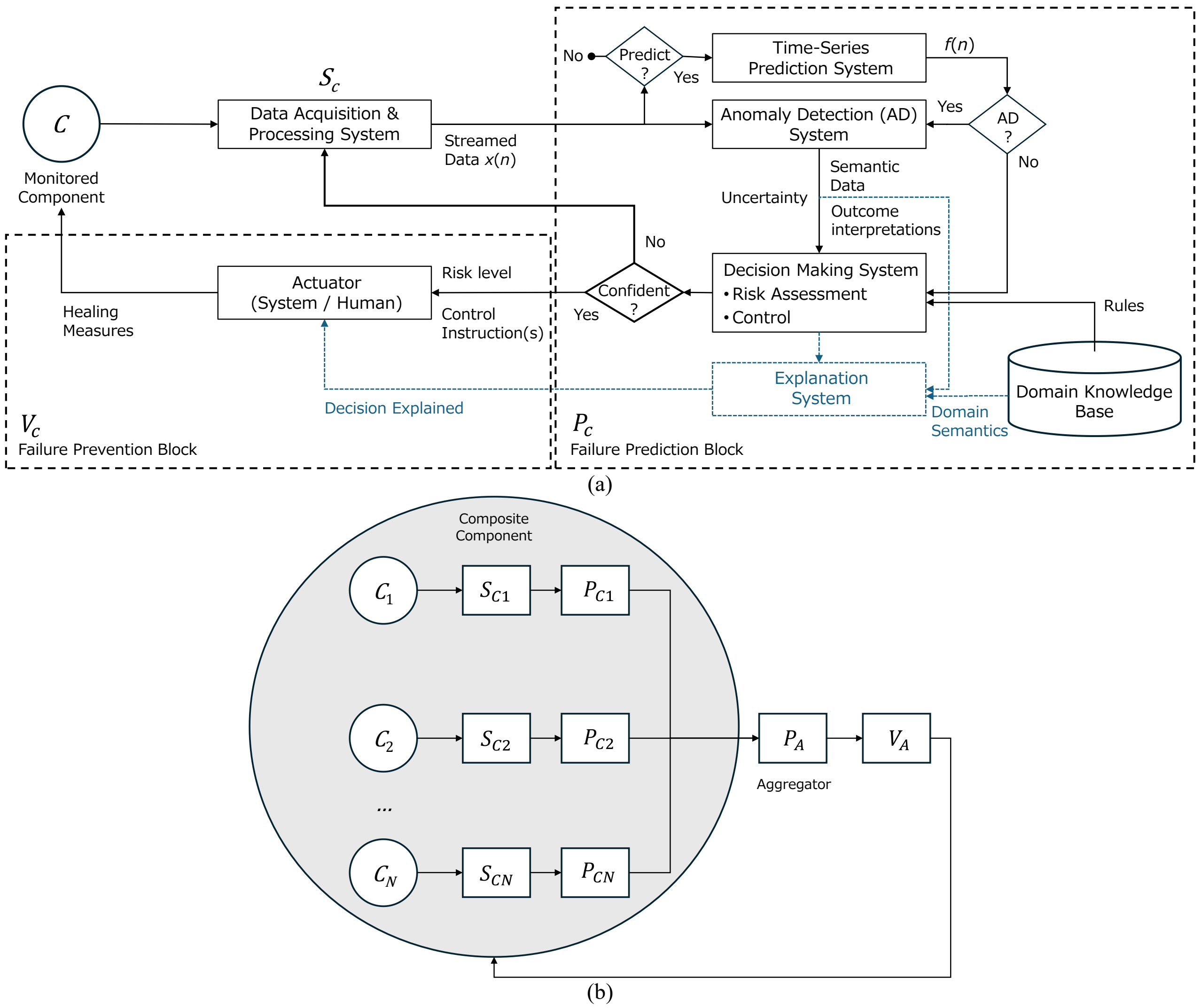

**Fig. 1** Overview of the proposed conceptual architecture. (a) Monitoring of a single component of a sea vessel for failure prediction and prevention; (b) monitoring of a composite sea vessel component, considered as a system of systems.

The failure of a system of systems (Fig. 1(b)), which can be a composite system of a sea vessel or the entire vessel, can be predicted by aggregating the outputs of multiple failure prediction blocks. These aggregates are then fused to provide a risk state of the system that expresses an ensemble of the confidence per subsystem. Subsequently, the output of the aggregator Failure Prediction Block ($P_c$) can inform a failure prevention block to take appropriate healing measures.

To make this concept more concrete, an example of the proposed architecture for the main engine of a vessel, is illustrated in Fig. 2. In this scenario, the main engine, which is considered a monitored component, has pre-installed sensors to monitor critical parameters, such as temperature, pressure, and vibration. The DAQ process involves streamed data derived from the engine component, including the cylinder ($T_c(t)$), exhaust temperatures ($T_e(t)$), and pressure $P(t)$. These data streams are the input of the $P_c$ (Fig. 2) that commences a forecasting procedure and the AD subsystem. The AD system identifies an out-of-distribution value in the cylinder head temperature data stream $T_c(t)$. When combined with the time-series system forecast of a further increase in temperature, these signals indicate a high probability of overheating in the main cylinder. The decision process receives expert-based domain knowledge and the AD System output providing possible risks, such as smoke in the engine room, cylinder and engine damage risk, along with a control recommendation to issue an alert to the chief engineer. A high confidence value triggers an incident of high risk with an "inspect engine" flag fed directly to the Failure Prevention Block $V_c$, where the actuator can see the possible failure cause and commence repairs.

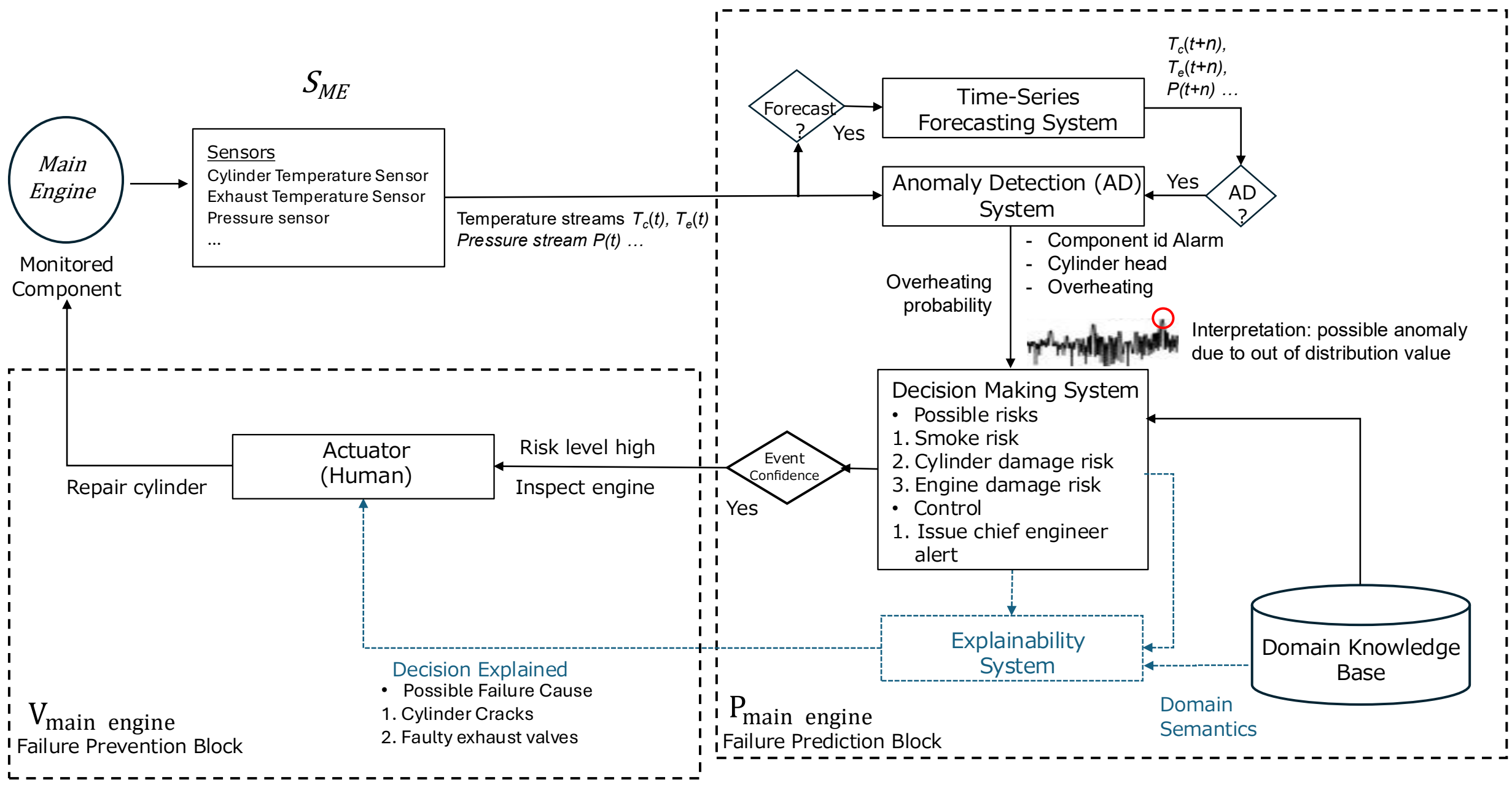

**Fig. 2** Integration of the main engine component into the proposed architecture.

Let us consider another scenario in the context of cybersecurity, simulating an attack on the vessel network that disrupts navigation [55]. The DAQ $S_{network}$ starts monitoring the radio frequency receivers (RF, VHF) that oversee the updating of navigation states. Following the rationale presented in the example in Fig. 2, the relevant position streams $P_{GPS}(t)$ and $P_{AIS}(t)$, and the network traffic stream $N(t)$ are first being fed to the $P_C$ for forecasting and AD. The AD system predicts a positional malfunction probability in conjunction with the semantic information of a GPS erroneous position and unusual network traffic that is interpreted as an out-of-distribution value. All these, combined with expert-based domain knowledge, provide valuable information to the decision-making system which uses it to identify potential risks, such as navigation discrepancies and a compromised network. A control recommendation signal for an officer alert, as well as an event confidence are the outputs of the decision-making system and the $P_C$ block. An incident activates the Failure Prevention Block $V_C$ that informs the actuator for an immediate inspection of the navigation systems with probable failure causes from the explainability system; namely, GPS spoofing (an attack that attempts to deceive a receiver with counterfeit signals)/jamming (an act of interference that paralyzes a receiver) [56] or faulty navigation systems (Global Navigation Satellite System (GNSS)).

In addition, it is important to demonstrate the multi-component facet of the architecture that encompasses the decision-making part of the conceptual framework. Within the complex network of systems-of-systems present on a vessel, a component that starts to show signs of failure might signify or infer that another component has already failed, especially in the case of components that are not observable. Fig. 3 illustrates an instance of the prediction-prevention closed-loop system, for a scenario where multiple components are monitored, such as the main engine, navigation readings, and network traffic. In this example, Failure Prediction Blocks, such as the engine PC1, the navigation PC2, and the network PC3, output low, medium, and high event confidence levels, respectively. The two latter components flag a component incident and, via an aggregated Failure Prediction Block $P_A$ , an operational incident. The reported incident is interpreted as a possible GPS spoofing/jamming situation, evidenced by navigation discrepancies or unusual traffic in the RF spectrum. Subsequently, the Failure Prevention Block $V_A$ receives the control recommendations to provide actuations that include system resets, network recalibrations, or even operational halt in shallow waters to avoid critical navigation hazards.

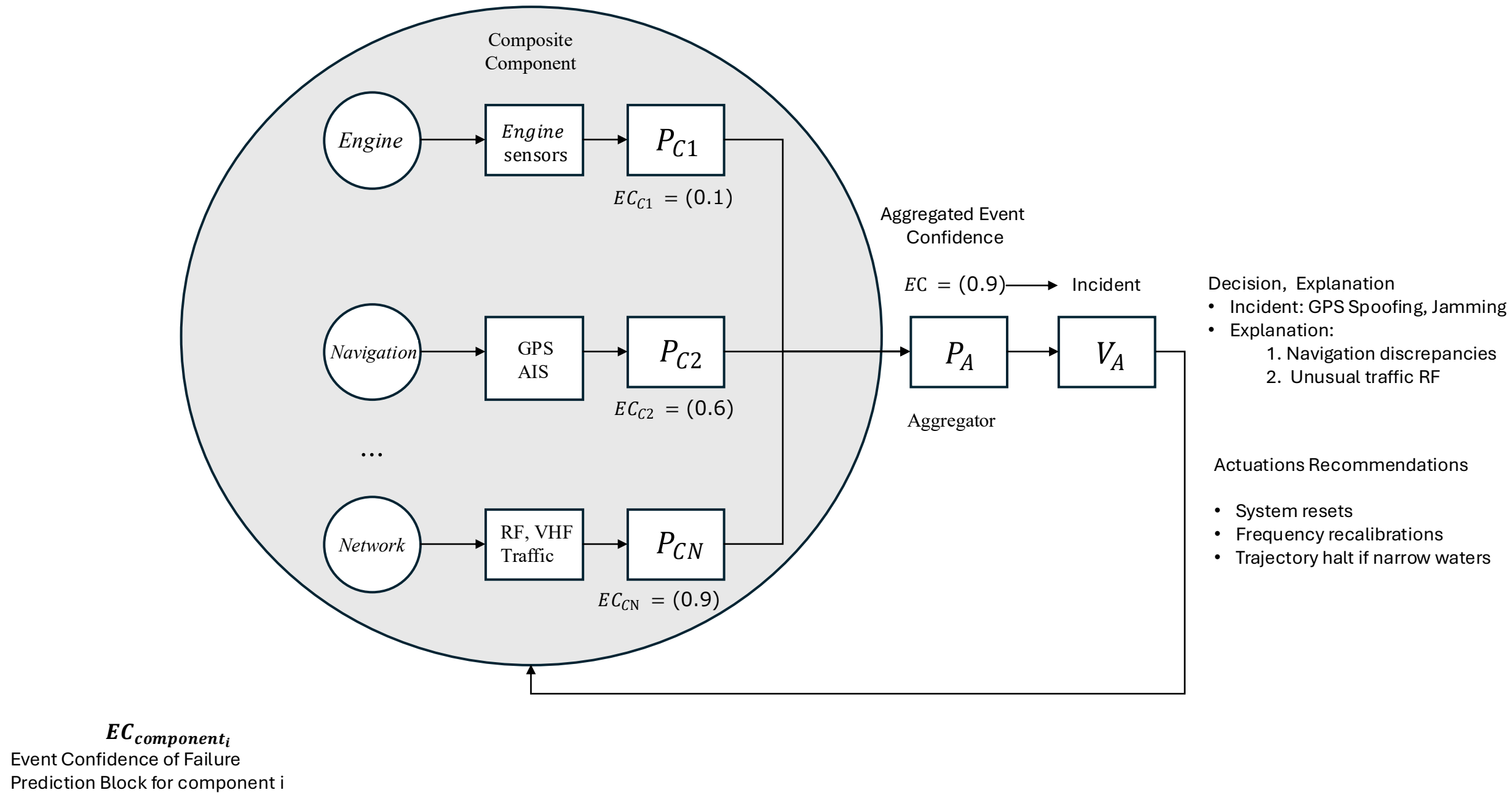

**Fig. 3** The multi-component prediction-prevention loop from the proposed architecture.

Overall, the proposed closed-loop system can provide extensive coverage, proactively predicting, detecting, and preventing unexpected vessel component behaviors, mitigating the operational costs and hazards. The above examples demonstrated that the proposed architecture can be easily adapted to the needs of each problem under consideration. It has the ability to combine multiple data collection channels and IoT sensor streams, and in combination with expert knowledge, it can provide interpretable outputs for predicting and detecting anomalies. Furthermore, based on the resulting predictions, it can alert personnel to potential failures that will occur, and take appropriate measures to prevent and/or limit the size of the failure.

In the following sections, we present a comprehensive review of recent studies related to the various modules and components of the proposed architecture. This overview highlights state-of-the-art methods and recent advancements that have been implemented in the maritime domain.

# 4 Data Acquisition and Processing

In PdM systems, sensors serve as the primary data collectors. Sensors are strategically installed on equipment and machinery to continuously monitor a diverse range of parameters, including temperature, pressure, vibration, and electrical current. The data captured by these sensors provide real-time insights into the operational health of the equipment, enabling subsequent PdM analyses. In maritime applications, the scope of data collection involves not only direct equipment parameters but also broader operational and environmental data, *e.g.,* vessel performance logs, Automatic Identification System (AIS) data, prevailing weather conditions, sea currents, and even marine life patterns. More specific examples from actual vessel operations include cylinder and exhaust gas temperatures, coolant temperatures, and cylinder pressures, all collected from various sensors integrated into the engine and machinery of the ship [57].

Raw data obtained directly from sensors frequently contain noise, inconsistencies, and missing values, requiring robust data preprocessing steps before actual analysis. This critical initial phase involves data cleaning to remove faulty readings, normalization to standardize data scales, and sophisticated handling of missing data points. It may also involve data transformation and feature extraction to derive informative representations, as well as data augmentation techniques to increase data diversity and improve model robustness. These steps constitute the backbone of any data preprocessing pipeline to ensure high-quality and high-integrity data for accurate PdM modeling. Furthermore, for systems integrating data from dissimilar sources, *e.g.,* Supervisory Control and Data Acquisition (SCADA), meteorological data, and accelerometers, careful data synchronization and interpolation are crucial to create cohesive and usable datasets [58]. The quality, completeness, and consistency of the input data, along with the rigor of preprocessing, directly influence the performance and reliability of downstream AI/ML models. This makes DAQ and preprocessing a crucial step, which is often underestimated. The inherent challenges of data quality and availability in harsh and dynamic marine conditions mean that even the most sophisticated AI models are only as effective as the data they are trained on [4]. This highlights the ongoing need for advancements in robust sensor technologies, the development and adoption of standardized data collection protocols across the industry, and the implementation of advanced data engineering pipelines to ensure information integrity and utility for trustworthy AI-driven decisions [59].

Vibration analysis is the principal methodology for SHM and fault diagnosis, especially for rotating machinery common in maritime environments [3]. This technique relies on the observation and detailed analysis of the unique vibration characteristics of a structure, which includes critical parameters of the vibration modes, such as the frequencies of the structure, the mode shapes, and the damping ratios. Advanced signal processing techniques are crucial for transforming raw vibration data into meaningful features [60]. Commonly employed techniques include time-domain analysis, frequency-domain analysis, time-frequency analysis, and order analysis/tracking. The aim of these signal processing techniques is to transform complex, high-dimensional raw sensor data into more manageable, low-dimensional feature vectors. This simplification facilitates easier data matching and comparison, making them suitable for subsequent AI-mediated analysis. The selection of an appropriate signal processing technique is not trivial, as it must align with the characteristics of the data and the target application. It directly affects the interpretability of the extracted features, since different techniques emphasize different aspects of the signal and determine how easily these can be related to the underlying physical behavior, as well as the overall effectiveness of the subsequent AI models. This is consistent with the inherent challenges associated with complex, non-stationary data and the significant environmental variability characteristics of maritime operations. AI/ML model efficacy for fault diagnosis and prediction depends highly on extracted feature quality and relevance. Misuse of signal processing methods can lead to misleading feature generation, causing AI models to learn false correlations or miss genuine indicators of impending failure, thereby compromising the reliability of the entire predictive system. This also highlights the importance of domain expertise and engineering judgment in both feature selection and data interpretation phases. Simion *et al.* [2] proposed an ML algorithm for evaluating onboard systems performance in maritime transport. They analyzed sensor data and mapped them to fault patterns to estimate functional states, integrating risk analysis tools like Fault Tree Analysis (FTA) and Failure Mode Effect Analysis with Criticality Analysis (FMECA). They employed the k-Nearest Neighbors (kNN) algorithm for real-time fault diagnosis and structured the algorithm with five operational modules. Their algorithm was validated through tests on a seawater cooling system from an oil tanker, using 20 operational parameters and 10 additional delta parameters. They reported that their algorithm reduced fault diagnosis time, enabled real-time monitoring, and was scalable. Hassan *et al.* [3] reviewed vibration-based condition monitoring studies. They found that PdM, using real-time equipment monitoring and data processing via sensors, predicts maintenance needs based on actual equipment status, thereby lowering costs and downtime. They also noted that it improves cost efficiency by shifting to on-demand maintenance and reduces unplanned downtime. Sehri *et al.* [61] proposed a framework for constructing and analyzing vibration-based spectrogram data. They started with bearing vibration data as an initial step towards creating a universal dataset for all machinery. They found improvements in model performance when pre-trained on bearing vibration data and fine-tuned on a smaller, domain-specific dataset, highlighting the potential to parallel the success of pre-trained NN for vibration analysis. Table 1 summarizes the sensor data types and the corresponding signal processing techniques for condition monitoring of different pieces of maritime equipment.

**Table 1** Sensor Data and Signal Processing Techniques for Maritime Equipment Condition Monitoring

| Sensor Data Type | Measured Parameters | Signal Processing Techniques | Typical Equipment/Application | References |
|---|---|---|---|---|
| Vibration | Oscillations, motions, modal parameters (natural frequencies, mode shapes, damping ratios) | Time-domain (SD, Pk-Pk, Skewness, Kurtosis, RMS), Frequency-domain (FFT, Spectral Analysis), Time-frequency (Wavelet Analysis), Order Analysis/Tracking | Engines, Motors, Pumps, Bearings, Gearboxes, Hull Structures, OWTs | [2], [62–65] |
| Temperature | Motor temperature (TEP), Cylinder/Exhaust Gas Temperatures, Coolant Temperatures | Data Preprocessing, Normalization, Feature Engineering | Engines, Motors, Pumps, General Marine Equipment | [2], [66–68] |
| Pressure | Cylinder Pressures, System Pressure | Data Preprocessing, Normalization, Feature Engineering | Engines, Pumps, Hydraulic Systems | [2], [66–69] |
| Current | Current Intensity (CI-T1, CI-T2, CI-T3), Winding Resistance (CR1, CR2, CR3) | Data Preprocessing, Normalization, Feature Engineering | Electric Motors, Generators | [2], [64], [65], [70] |
| Acoustic Signals | Noise signals | Statistical, Spectral, Time-frequency Analysis | Engines, Rotating Machinery | [2], [63], [68], [70] |
| Operational Data | SCADA parameters (pitch, power, RPM, wind speed, yaw, wind direction), METEO data (tidal level, wave height, sea water temperature, air pressure), Telemetry, Maintenance History, Machine Specifications, AIS data | Normalization, Synchronization, Interpolation, Feature Engineering | All Vessel Systems, Fleet Management, Route Optimization | [2], [62], [66], [67] |
| Image/Video | Thermal distribution (Infrared), visual structural condition, 2D time-frequency representations (converted from 1D signals) | Computer Vision (CNNs, ResNet), Signal-to-Image Conversion, Image Segmentation, Thermal Pattern Recognition | Marine Diesel Engines, Marine Centrifugal Fans, Rotating Machinery, Gearboxes | [68], [71–73] |

# 5 Time-Series Forecasting & Anomaly Detection in Maritime

Time-series prediction in the maritime sector primarily involves the use of advanced analytics and AI to forecast events, optimize operations, and enhance decision-making. Predictive intelligence in this industry helps anticipate delays, improve supply chain efficiency, ensure regulatory compliance, and reduce operating expenses by analyzing vast amounts of data such as vessel movements, weather conditions, and port operations [74].

## 5.1 Time-Series Forecasting

Within the framework of maritime operations and systems, it is often necessary to forecast a given operational state of a vessel within a given window, in order to reliably ensure safety and efficiency of a naval operation. Time-series forecasting for maritime systems methodologies can be summarized in two main approaches, namely simulation-based and data-driven approaches, including statistical and ML models.

**Simulation-based Approaches & Digital Twins**

Simulation-based approaches have long been employed in the maritime domain to model complex systems and evaluate operational performance, particularly in scenarios where real-world experimentation is impractical. Carotenuto *et al*. [75] proposed a simulation-based framework to evaluate maritime transport within an oil supply chain, where they model the full logistics process, from refinery to coastal depots, and use stochastic simulation to analyze how ship traffic affects inventory variability and operational decisions. Their work is clearly simulation-driven, focusing on "what-if" analysis to optimize supply chain performance under uncertainty. Onggo *et al*. [76] followed a similar modeling philosophy but shifts toward operational decision-making and optimization, using mathematical modeling and simulation techniques to evaluate maritime or transport system performance under uncertainty, again relying heavily on simulated scenarios rather than real-time systems. Han *et al*. [77] proposed a temporal multi-agent simulation framework for ship navigation in port environments, where vessel movements, waterways, and port infrastructure are modeled as interacting agents to simulate ship entry, exit, and traffic flow dynamics; their model was used to evaluate port capacity and operational efficiency under realistic conditions. Pham *et al*. [78] introduced ShipNaviSim, a data-driven maritime traffic simulator that learns vessel behavior from AIS data and reproduces realistic multi-ship interactions, enabling the generation of complex traffic scenarios and supporting navigation system testing. Baldauf *et al*. [79] investigated full-mission maritime simulation environments combining ship-handling and vessel traffic service (VTS) simulators, demonstrating how integrated simulation scenarios can replicate complex operational conditions and improve safety analysis in congested waterways. While the aforementioned works rely on simulations to model maritime systems, evaluate operational scenarios, and analyze system behavior, they typically operate in an offline or scenario-driven manner without continuous interaction with real-world systems. This distinction leads to the concept of the Digital Twin (DT), which extends traditional simulation by enabling a dynamic, data-driven, virtual representation of a physical asset or system. A DT can be dynamically updated through the continuous exchange of information between the physical and virtual states. This model extends beyond mere static simulation by integrating real-time sensor data with historically collected records, thereby reflecting the actual status of the asset across its entire life cycle. Furthermore, this bidirectional mirroring enables not only real-time monitoring, diagnostics, and predictive analytics but also the evaluation of future behaviors, scenario forecasting, and the optimization of operational actions [80–82]. DTs excel in real-time monitoring, simulation, and data-driven insights, leading to improved operational efficiency, reduced costs, and enhanced safety. These dynamic models are designed to accurately replicate the real-time behavior of physical assets, providing immediate feedback and enabling simulation and forecasting scenarios. A series of other studies have also reviewed the emerging trend of DTs in the maritime domain [83–86].

Tsitsilonis *et al*. [87] presented a DT-enabled framework for the health assessment of marine diesel engines. The inputs included engine shop-test data, operational measurements, and corrected in-cylinder pressure measurements, which were used to build DTs representing both healthy and current engine conditions. The outputs were health and performance indicators, including Brake Specific Fuel Consumption (BSFC), peak in-cylinder pressure, Indicated Mean Effective Pressure (IMEP), and exhaust gas temperature differences, enabling the identification of underperforming cylinders and degradation across the engine operating envelope. The study shows that the DT can map engine degradation, support diagnostics, and improve lifecycle decision-making even when limited measurements are available. Taghavi and Perera [88] proposed a DT-type modeling framework for marine engine fuel consumption prediction. The inputs consist of operational engine data from an ocean-going vessel, organized into operating regions using Gaussian Mixture Model-based clustering. Local polynomial regression models are then developed for each region and combined through Multiple Model Adaptive Estimation (MMAE). The output is the predicted fuel consumption under given engine operating conditions, allowing the DT to adapt across different operating regimes and support energy-efficient ship operation. Jeon *et al*. [89] developed data-driven Prognostics and Health Management (PHM) models using datasets generated from a physics-based DT of a four-stroke marine engine with stochastic exhaust valve degradation. The inputs include simulated operating conditions, degradation profiles, and health-related engine variables generated by the DT. The outputs are constructed health indicators and their predictions, which support long-term prognosis and RUL-oriented assessment of engine components while addressing the limited availability of real-world degradation and anomaly data in marine machinery. Perera *et al*. [90] developed a Decision Support System (DSS) that leverages a DT framework for green ship operations and advances research in ship energy efficiency and emissions reduction, contributing to the transition toward net-zero emission vessels. The inputs include vessel operational scenarios, existing and alternative fuel options, emerging green technologies, and pilot-vessel data. The DT integrates multiple simulation capabilities, such as voyage optimization, Life Cycle Cost Analysis (LCCA), and Environmental Impact Assessment (EIA), with the scope of evaluating both existing and emerging green fuels and technologies. The outputs of the DT are the decision-support results from simulations such as voyage optimization, life cycle cost analysis, and environmental impact assessment, supporting the evaluation of emission reduction strategies and the transition toward net-zero vessels.

**Statistical methods & ML**

Advanced time-series analysis and forecasting is vital for optimizing maritime operations with growing success in accurately anticipating vessel activity, traffic flows, and environmental conditions, driving major gains in efficiency, safety, and informed operational planning across the industry. Patmanidis *et al*. [91] proposed an AIS-based maritime surveillance framework for vessel route estimation and abnormal behavior detection. Their approach uses linear filtering through ARMA/ARX models to forecast future vessel positions from past latitude and longitude measurements, after resampling AIS data into constant time intervals. In addition to trajectory forecasting, they define alert criteria related to abnormal route deviation, sudden course changes, excessive speed, inconsistencies between position and velocity, and missing AIS transmissions. Ramin *et al*. [92] focused on marine traffic density forecasting in the Straits of Malacca using AIS-derived vessel counts. Their study compares classical quantitative forecasting methods, including moving average, weighted moving average, exponential smoothing, and multiple regression with seasonal variables. The results show that exponential smoothing achieved the best performance for the task of short-term traffic density prediction. Li *et al*. [93] proposed a similarity grouping-guided NN framework for maritime time-series prediction, targeting complex, nonlinear, and non-stationary series, such as port cargo throughput and vessel traffic flow. Their method first decomposes the original signal into high- and low-frequency components; the former are predicted directly with traditional neural networks, while the latter are segmented and grouped using dynamic time warping (DTW) to create more suitable training samples. The final prediction is obtained by recombining the predicted components, and experiments revealed improved prediction accuracy and robustness compared to conventional NN-based approaches. Liu and Ma [94] proposed an Attention-LSTM model for ship navigation behavior prediction using AIS data. Their work first addresses AIS data quality through navigation data extraction, abnormal data processing, and missing-value interpolation, then uses dynamic navigation variables, *e.g.,* longitude, latitude, speed over ground, course over ground, heading, and time increment, as inputs to predict future ship longitude, latitude, heading, and speed. By introducing an attention mechanism into the LSTM structure, the model better captures important spatiotemporal features in the AIS sequence, and simulation-based experiments demonstrate improved accuracy and robustness for ship navigation prediction. Zhang *et al*. [95] provided an extensive review on state-of-the-art prediction support systems for vessel trajectory prediction in maritime transportation, emphasizing recent advancements, especially DL approaches. Predicting trajectories utilize data sources, such as AIS, Vessel Monitoring Systems (VMS), radar, satellite images, or electronic navigational charts and scenarios that can be categorized. They concluded that, although state-of-the-art methods, such as Kalman filters and NNs (GANs, Transformers) can provide procedures with detection capabilities, hybrid models that combine multiple techniques are more efficient. Furthermore, Li *et al*. [96] applied ML-based forecasting modeling to naval propulsion systems, specifically Diesel engines, by developing data-driven models that forecast temperature variations indicative of potential faults. Their research delivers an approach that enhances safety, maintenance effectiveness, and operational cost management. In their study, they tested a range of non-probabilistic and ML techniques including Autoregressive Distributed Lag (ARDL), stepwise regression, and various NNs to model both linear and nonlinear relationships in time-series sensor data, focusing on temperature measurements and forecasts. To validate their approach, they standardized a data preprocessing technique with various autonomous feature selection methods and performed extensive tests on simulated data promoting model efficiency and accuracy.

Prediction modeling and time-series forecasting play an important role in various aspects of data-centric control processes within the maritime domain, especially when integrated as a pillar component subsystem in systems that deal with detecting abnormal states. These deviations from normal patterns include a collision trajectory, a degraded engine cylinder, or a compromised network that falsify important readings and disrupt operational safety. Thus, it is vital that these states can be predicted as early as possible to avoid safety risks, high repair costs, and even operational halts. Time-series forecasting and AD in the maritime domain are closely linked, as understanding normal operational patterns is closely related to being able to forecast them and is a prerequisite for identifying abnormal or risky behavior; therefore, the next chapter will explore this connection and its practical implications in greater depth.

## 5.2 Anomaly detection in Maritime

In the maritime sector, the integration of wireless communications and sensing technologies is becoming standardized and the ever-increasing design of systems-of-systems that require the processing of vast amounts of data in an autonomous manner drive the research in the domain to improve maritime services [97]. Naval transportation systems are envisioned to rely on real-time monitoring and automated optimization methods to achieve ship path optimization, energy consumption reduction, and speed optimization [98]. Furthermore, PdM [99] and Network Security (SecOps) [100] provide systems that can guarantee the safety of personnel and the avoidance of disruption in maritime operations. It is evident that an early detection of conflict can provide enough time to take appropriate actions before potential problems occur. This section describes application case studies where maritime operators have integrated forecasting or AD methods in various aspects of the operational lifecycle of a vessel.

### 5.2.1 Maritime Operations Monitoring

With more than a million ships tracked globally, manually detecting suspicious ship behavior is impractical. AD and forecasting methods have been proven effective when integrated with traditional maritime monitoring tools to enhance operations monitoring stability and safety in the real or cyber domain.

**Vessel trajectory monitoring and Collision Avoidance**

A common application of AD in the maritime domain is to track the path of a vessel and define anomalous deviations from the relegated path. AIS has been introduced in maritime operations to enhance sea traffic safety. AIS messages are broadcast to nearby vessels and include details such as the identity, location, speed, and course of the sender. This system helps prevent collisions and improves onboard situational awareness. Consequently, a variety of automated methods for identifying anomalies in maritime AIS data have been developed [74]. These detectors leverage the somewhat predictable nature of vessel movements, especially within confined areas, and the physical or regulatory limits on their behaviors. They typically build models of standard vessel traffic patterns and flag deviations or irregularities as anomalies, which could indicate incidents such as accidents or illicit activities. Focusing on spatial AD in maritime settings, five primary types of abnormal behavior derived from AIS data can be identified, as illustrated in 4. In each subfigure of Fig. 4, the arrows and data marked in orange indicate a potential deviation from the normally expected patterns. In the ML domain, Fig. 4. (a) (c) (e) are considered contextual anomalies, Fig. 4 (b) is a point anomaly, and (d) is a collective anomaly.

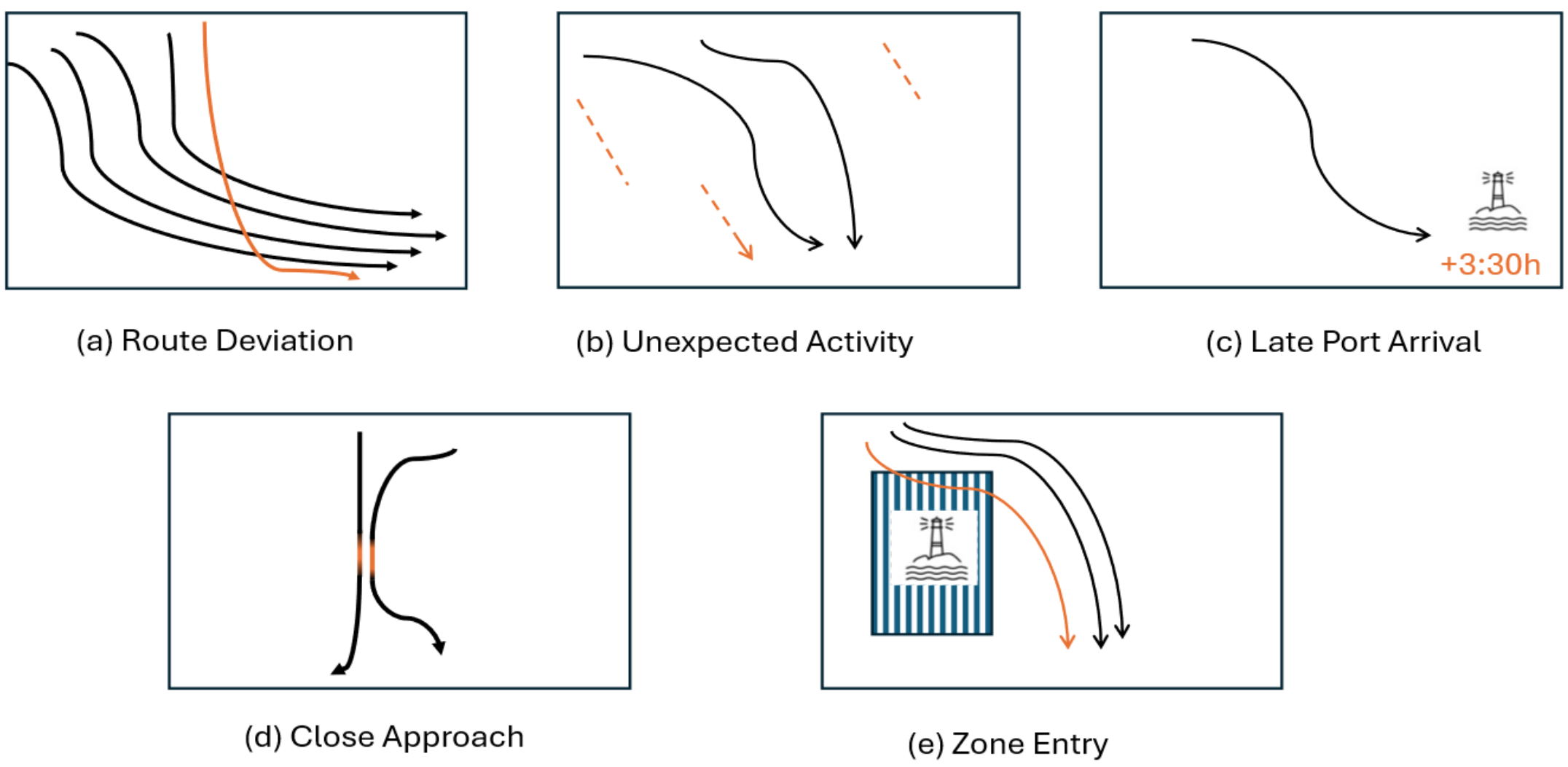

**Fig. 4** These general five AIS anomaly types are used in [101] to classify the capabilities of published anomaly detectors. (a) Route deviation; (b) unexpected vessel activity; (c) late port arrival; (d) two vessels closely approaching; (e) illegal zone entry.

Wolsing *et al*. [100] presented a thorough review on AD in maritime AIS tracks where they compared several state-of-the-art operational and theoretical methodologies. The deterministic operational approach for collision RA uses parameters such as relative distance, speed, course, and compliance with the International Regulations for Preventing Collisions at Sea, *i.e.,* Convention on the International Regulations for Preventing Collisions at Sea (COLREG). The models reviewed in that study enhance traditional, deterministic proximity measures and incorporate a ship path optimization model based on a network-based or mixed-integer nonlinear programming approach that balances safety (by minimizing collision risk) with operational efficiency (by reducing sailing time), while accounting for vessel behaviors and geographical constraints. These methodologies contribute to the development of intelligent navigation systems that minimize human error and enhance traffic management. The main current limitation is the scarcity of large publicly available datasets; hence, much research has relied on restricted data from military or defense sources. Nonetheless, AD in ship trajectories remains a vital area for future research to safeguard maritime operations.

**Cybersecurity and Network Monitoring**

A domain where the application of AD has become crucial is that of cybersecurity and network monitoring. The rise of Cyber-Physical Systems has introduced new security challenges. In recent years, naval systems have increasingly adopted Intrusion Detection Systems (IDS) to protect Programmable Logic Controllers (PLC) and Industrial Control Systems (ICS) due to plenty of vulnerabilities. The challenge of managing information system security on ships can be summarized in two key points: first, SCADA systems and industrial networks are designed for long operational lifespans onboard, typically 15 to 20 years, which makes them difficult to fix and upgrade; second, the programming practices used in these systems often do not prioritize safety [102]. The vulnerabilities of industrial systems no longer need to be demonstrated, including issues such as insecure development practices, weak protection access controls, management information systems that are not hermetic, outdated protocols, and minimal system supervision. Consequently, industrial control systems in critical infrastructure, such as those on cargo ships, are highly susceptible to cyber-attacks. In particular, Programmable Logic Controllers (PLCs) are vulnerable to various attacks including recognition, authentication bypass, CPU stop and start, brute-force attacks, command injection and response, Denial of Service (DoS), and firmware tampering. Boudehenn *et al*. [103] further investigated security vulnerabilities in maritime network systems and proposed an AD algorithm based on the multi-resolution Teager-Kaiser energy operator. This method effectively identifies injected anomalies, thereby enhancing alert management onboard ships. Their results demonstrate that monitoring sensors and actuator alerts is essential for maintaining cybersecurity in naval systems. Experiments also highlighted the significant effect attacks can have on a ship propulsion management system.

### 5.2.2 Predictive Maintenance

Predictive maintenance, a domain where AD has gained significant traction, has the main goal to proactively predict equipment failures using data from various sensors embedded in a vessel's operating system and components in order to prevent both unexpected failures and premature replacements, reducing both planned and unplanned downtime, lowering maintenance costs, and eliminating unnecessary maintenance activities.

**Marine Engines and Propulsion Systems**

Fault diagnostics in internal combustion engines (ICEs) are vital for optimal operation and the prevention of costly breakdowns in marine vessels, since the engine constitutes the determining component of a ship. An engine failure can render a ship inoperable or severely restrict its functionality. Srinivaas *et al*. [104] conducted a review study on methodologies for ICE fault detection, including model-based and data-driven approaches. They reported that advanced DL techniques, e.g., DNNs, one-dimensional (1D)-CNNs, Transformers, and hybrid Transformer-DNN models, demonstrated superior performance in fault detection compared to traditional ML methods. Pajak *et al*. [59] proposed a novel system-oriented method of reliability state identification for marine engines. They analyzed vibration and noise signals collected from engine cylinders using supervised ML, with a key novelty being the application of data augmentation and SVM classifier implementation. Their method was robust even with poor-quality or limited/incomplete learning datasets, achieving high-quality identification of reliability states and fault types. The shift towards data-driven and DL approaches for engine diagnostics reflects the growing complexity of modern marine engines and the inherent limitations of traditional model-based methods in capturing their intricate, non-linear degradation patterns. This indicates that while physics-based models are accurate when degradation mechanisms are well-understood, they struggle with the dynamic operating conditions and complex interactions within modern marine engines. Furthermore, RUL prediction for turbofan engines has been a significant focus area, often utilizing publicly available time-series datasets such as C-MAPSS [105]. LSTM networks, known for their ability to process sequential data, have also been successfully employed to estimate the RUL of diesel engines [4].

**Shipboard Motors and Pumps**

PdM is a critical strategy for ensuring the reliability and operational efficiency of industrial systems, including essential shipboard components such as electric motors and pumps. Supervised learning models have been actively employed to diagnose the condition of electric motors. For example, Baradaran [106] developed a predictive model for determining the condition of electric motors. A dataset of motor temperature, current intensity, winding resistance, and sound condition was utilized, and various supervised learning models were trained and evaluated, including Naive Bayes, SVM, Logistic Regression, k-NN, Random Forest, XGBoost, LightGBM, and CatBoost. The study reported that CatBoost achieved the highest accuracy (92.86%), demonstrating superior performance due to its efficient handling of categorical data and advanced regularization techniques. For shipboard pumps and other operational systems on vessels, robust AD capabilities are paramount [57]. Moghadam *et al*. [107] investigated vessel operational AD using semi-supervised DL models augmented with lightweight interpretable surrogate models. They utilized time series records from IoT datasets, including cylinder and exhaust gas temperatures, coolant temperatures, and cylinder pressures from a sensorized vessel. Data enrichment and validation were achieved by generating synthetic outliers through data degradation and augmentation with a Transformer backbone. An improvement in performance of 17.23% was reported compared to existing state-of-the-art methods across five publicly available datasets and actual maritime operational data. The LSTM autoencoder exhibited high precision and recall (over 80% and 90%, respectively) in detection, while lightweight interpretable models (Random Forest, Decision Tree) provided helpful interpretability for engineers.

**Bearings and Other Rotating Machinery**

Rolling bearings are fundamental components in a wide array of rotating machinery, and their susceptibility to faults under high loads and challenging operational environments makes their condition monitoring crucial. Timely intervention for bearing faults is essential to prevent cascading failures that can lead to critical system breakdowns. Eddai *et al*. [60] investigated the application of ML algorithms for bearing fault detection. They employed a systematic approach including data preprocessing (filtering, signal decomposition), feature extraction (time- and frequency-domain features), and fault classification using SVM, kNN, Decision Tree, and Naive Bayes algorithms. They achieved an aggregate classification precision of 97.8% across all fault categories, with the SVM algorithm demonstrating exceptional performance at 99.2% accuracy in inner-race fault identification. Brusa *et al*. [108] investigated the capabilities of SHapley Additive exPlanation (SHAP) for feature importance in ML models for industrial condition monitoring, specifically for medium-sized bearings. They applied SHAP to explain diagnosis outcomes of SVM and kNN models using vibration data. They found that the skewness and shape factor of the vibration signals were the most important features for classifying bearing defect conditions, while diagnostic accuracy was not significantly affected when the number of features was reduced based on SHAP values. The importance of explainability has been addressed by Nor *et al*. [109], who systematically reviewed XAI applications in PHM for industrial assets. They found a growing acceptance of XAI in PHM, noting that XAI offers dual advantages by both executing PHM tasks and explaining diagnostic and AD activities, without negatively affecting PHM performance. The identified key challenges include lack of explanation evaluation metrics, insufficient human involvement in explanation evaluation, and inadequate uncertainty management. Moreover, Mey and Neufeld [110] investigated the application of XAI algorithms (GradCAM, LRP, and a modified LIME) to Convolutional Neural Networks (CNNs) for vibration-based condition monitoring. They transformed vibration data into Frequency-RPM Maps and Order-RPM Maps for CNN input. Each investigated XAI algorithm was relatively successful in highlighting class-specific features, providing additional insights into vibration signals, *e.g.,* revealing fault-specific vibrational modes. By understanding which features (*e.g.,* skewness, shape factor) are most discriminative, researchers and engineers can optimize data collection and preprocessing efforts, potentially reducing data dimensionality and computational load while maintaining or even enhancing diagnostic accuracy.

**Applications in SHM of Offshore and Marine Structures**

Vibration-based SHM is a widely adopted strategy for ensuring structural integrity and optimizing the lifespan of large infrastructure, including critical marine installations such as offshore wind turbines (OWTs). While OWTs are not ships, they present analogous challenges due to their constant exposure to harsh marine environments and the complex effect of environmental and operational variability on their structural response [111], [112]. Weil *et al*. [58] introduced a novel vibration-based SHM methodology that integrates ML and Uncertainty Quantification (UQ) for OWTs. They used Automated Operational Modal Analysis (OMA), DBSCAN for clustering, and a "human-in-the-loop" approach for data selection. Data were pre-processed using circular encoding and min-max normalization, feature selection was performed using RFECV with XGBoost, and various ML models were compared. CatBoost was eventually selected for its built-in UQ capabilities (virtual ensembles). They developed a "smart tracking" method based on ML predictions and UQ thresholds and used control charts for decision-making. They reported significant improvements in SHM processes for OWTs, particularly in natural frequency tracking and AD, effectively removing rotor harmonic interference and showing improved AD in synthetic datasets. Expanding on this need for robust diagnostics, other researchers have employed unsupervised deep learning to tackle complex monitoring tasks. For example, Bel-Hadj and Weijtjens [113] utilized deep autoencoders on the spectrograms of acceleration signals to create an agnostic anomaly detection system for OWTs, enabling the early identification of structural deviations without relying on extensive feature engineering.

# 6 Decision-Making and Risk Assessment

## 6.1 Risk Factors

Maritime RA can be assessed based on four risk factors related to the a) static characteristics of the ships, b) meteorological conditions, c) ship dynamics, and d) their position compared to shipping lanes [49], [114]. The static risk factor is a numerical estimate of the inherent level of risk associated with a ship, based on fixed characteristics. These ship-related factors are affected by elements such as the year of construction (recent or old), flag, size, service history, proper maintenance, and the design or type of ship (*e.g.,* tankers, container ships). In addition, the experience and training of the crew serve as vital static indicators, as less experienced personnel are more prone to errors and improper handling of risks. Overall, this static risk can be calculated using a fuzzy logic-based model to manage uncertainty in maritime safety assessments [114].

Unlike static characteristics, environmental and meteorological factors are largely dynamic and require constant monitoring. These include weather conditions, such as strong winds or storms, which significantly increase the risk of accidents. Furthermore, visibility, which is affected by fog, rain, or night operations, can severely impact navigation and increase the risks of collision or malfunction. Closely related to these variable conditions are ship dynamics, which refer to characteristics such as the speed and direction of the vessel. These dynamic factors directly affect the likelihood of operational accidents, such as collisions and grounding.

Another critical element of assessing situational risk is the position of the ship in relation to established shipping lanes. Shipping lanes are defined routes intended to organize maritime traffic and reduce the likelihood of encounters on the high seas. A ship operating outside or near the edges of these lanes, especially in high-density areas or traffic separation schemes, may face an increased risk of collision or create navigational conflicts. Similarly, vessels sailing against the recommended direction in a one-way lane or crossing lanes at sharp angles pose a significantly higher risk.

Beyond these fundamental operational and navigational factors, external issues can further complicate maritime operations, particularly piracy and security threats. Attacks from outside are possible, with ships potentially being diverted from their planned routes by interested external forces. In addition, although previously considered a marginal problem for maritime transport, cyber risk is now considered one of the greatest threats to the international economic environment. These risks require a prompt response, proper organization, and appropriate measures to address them. To manage these overarching operational and security challenges, maritime conventions regulated internationally by the International Maritime Organization (IMO), such as SOLAS (International Convention for the Safety of Life at Sea), MARPOL (International Convention for the Prevention of Pollution from Ships), STCW (International Convention on Standards of Training, Certification and Watchkeeping for Seafarers), and COLREGs , have been continuously developed and amended over time to ensure maritime safety.

## 6.2 Maritime Decision-Making Methods

Multi-Criteria Decision Making (MCDM) methods, including the Analytic Hierarchy Process (AHP) and Technique for Order of Preference by Similarity to Ideal Solution (TOPSIS) were examined and presented in [23] for route selection and emergency decision-making tasks, including ice prediction and unexpected weather conditions. In [115], Arctic waters were analyzed by identifying accident-prone areas based on historical data from 2010 to 2020, and a directed-weighted emergency port network was presented; this network incorporated both emergency rescue time and port emergency capacity to evaluate and rank candidate ports for responding to maritime accidents. Accident regions were clustered using K-means, and potential ports were assessed through a multi-criteria decision-making approach, combining historical data and entropy-weighted TOPSIS. A variety of marine accident consequence scenarios were predicted in [116] using a methodology that was based on the Classification and Regression Tree (CART) and Random Forest models, optimized through Random Search and Bayesian Optimization. The BO-RF (Bayesian optimized Random Forest) model achieved the highest prediction accuracy, compared to other state-of-the-art approaches. A feature importance analysis conducted in this study identified the number of people involved as the top influencing factor. A decision-making support approach for shipping safety was developed by integrating neurophysiological and subjective data through functional connectivity analysis, based on functional Near-Infrared Spectroscopy (fNIRS) [117]. The proposed methodology enabled the detection of different workload levels, and the assessment of human reliability in complex scenarios, comparing brain activity patterns.

Collision RA based on the vulnerability of marine accidents was presented in [118] using fuzzy logic. Specifically, fuzzy logic was employed to reason basic collision risks using Distance to Closest Point of Approach (DCPA) and Time of Closest Point of Approach (TCPA) and the degree of vulnerability in the specific coastal waterways. AHP was also used to obtain the integration of vulnerabilities. In [119], a wide variety of decision-making methodologies were proposed and applied to various challenges within the maritime domain, using specific case-studies. AHP and its extension, based on fuzzy logic, called Fuzzy AHP (FAHP), were utilized to support strategic decision-making in maritime logistics. The AHP method was applied to structure complex decision problems hierarchically and derive priorities through pairwise comparisons. To better address uncertainty and vagueness in expert opinions, FAHP was incorporated, thus enhancing the robustness of the resulting decisions.

Furthermore, a hybrid fuzzy-Delphi-TOPSIS model was proposed for bunkering port selection. More specifically, the Delphi method was utilized to select the opinion of experts and fuzzy logic was integrated, aiming to linguistically characterize the uncertainties. The TOPSIS method was applied to rank candidate ports in terms of proximity to the ideal solution. Heuristic algorithms, including dynamic programming, simulated annealing, and Non-dominated Sorting Genetic Algorithm II (NSGA-II), were applied to optimize container terminal operations, including container loading, and gate resource allocation. A combination of the Analytic Network Process (ANP) and Interpretive Structural Modeling (ISM) was also utilized to evaluate environmental sustainability in ships. In particular, ANP was used to model the relations among various factors influencing green shipping practices, while ISM contributed to the interpretation of these relationships. Moreover, a Fuzzy Evidential Reasoning (ER) approach was utilized for the selection of vessels under uncertain operating conditions. This methodology enabled the fusion of both qualitative and quantitative criteria and was particularly effective in scenarios involving incomplete or imprecise information. A hybrid model combining the Artificial Potential Field (APF) method with Evidential Reasoning (ER) was presented for the assessment of maritime collision risks. The APF component modeled dynamic risk zones based on the relative motion and proximity of vessels, while the ER component processed various sources of uncertain data to predict the risk of collision. The Dempster's rule of combination was also used aiming to aggregate the results from the different sources used. A Stochastic UTA (Utility Additive) method has also been employed to evaluate the risk of ship-to-ship transfer operations, aiming to address uncertainties among stakeholders. A human factors analysis in maritime accidents was presented in [120], using a dynamic fuzzy Bayesian network. The proposed methodology incorporated intuitionistic fuzzy set theory and Bayesian networks to deal with uncertainty. In addition, the proposed approach was used to model various sand carrier accident scenarios from China, while identifying the most representative human factors that caused the maritime accidents. Maritime accident modeling focusing on ship collisions was presented in [121], using an extended Bayesian network framework with interval probabilities to explicitly quantify uncertainty evoked from expert judgments.

A few recent studies have focused on the interpretability of black-box models in the maritime domain, using post-hoc approaches, including SHAP. All these studies developed posterior explanatory models, *i.e.,* they first built a black-box ML model and then developed an additive white-box model based on SHAP to interpret the results of the black-box model. In detail, SHAP assigns an important value to each feature value for all the predictions [122]. By aggregating the important values of all feature values in a sample and the mean value of the predicted target in the training dataset, SHAP develops an additive white-box model to explain the predicted values of each sample. Therefore, SHAP provides insights into how the predicted values are obtained in a black-box model and thus is suitable for ex-post explanations. Additive white-box models based on SHAP and Local Interpretable Model-agnostic Explanations (LIME) for predicting maritime accidents were utilized in [123], while providing interpretations of their predictions. In [124], SHAP was utilized to interpret the outcomes of tree-based ML models in predicting maritime accidents, whereas in [125] SHAP was used to explore the internal mechanisms for classifying the types of ships. In [126], the factors affecting ship detention were identified using SHAP, and in [127] SHAP was adopted to provide explanations for the port state control problem.

# 7 Challenges and Perspectives

The above review of methodologies aimed at exploring the aspects of a closed-loop system for failure detection and prevention in ships, covering issues related to sensor technologies, AD modeling, RA, and management. This work highlighted the research challenges that should be considered when developing systems for automatic damage detection on ships. These challenges can be summarized as follows:

(i) **Uncertainty, noise, and robustness**. Predictive models often operate under unknown or changing conditions, including unseen operating regimes, noisy sensor measurements, missing values, and shifting data distributions. These factors can reduce model reliability and hinder decision-making. Therefore, future methods should account for uncertainty and be robust to noise, out-of-distribution data, and changing operational profiles.

(ii) **Generalization across vessels and components.** Several models are trained on data from a specific vessel, engine, sensor setup, or operating condition. This limits their transferability to other maritime assets. Future work should focus on domain adaptation, transfer learning, and hybrid physics-informed approaches that improve generalization across different systems.

(iii) **Explainability & Trust**. In certain safety-critical areas, designing trustworthy systems that rely on black-box methods poses significant risks. To reduce these risks, explainable algorithms are essential as they offer clear insights into how the decision-making process works. Several AI models, especially those based on DL, are considered as “black boxes”, meaning that while they can make highly accurate predictions, it is often difficult to explain how and why they reach these specific conclusions. To address this challenge, the concept of XAI has been increasingly utilized and designed to explain the outcomes of AI models. XAI enables users to understand the processes behind AI predictions and decisions, providing insights into how models reach their conclusions. Furthermore, XAI has a great impact on naval equipment maintenance, where precise decisions are required, affecting equipment safety and reliability.

(iv) **Lack of accessible datasets.** Real maritime anomaly and failure datasets are rare since data are often proprietary, sensitive, or collected under industrial and military projects. This limits reproducibility and fair comparison between methods. Open benchmark datasets as well as synthetic datasets and simulation-based anomaly generation, are needed to support wider research and validation..

(v) **Operational evaluation and deployment.** Several studies evaluate models only through standard accuracy metrics, which may not reflect real maintenance needs. Future work should consider false alarms, missed detections, detection delay, computational cost, and real-time applicability. Human-in-the-loop decision support is also important, so that anomaly scores, forecasts, and explanations can be effectively used by operators and engineers.

# 8 Conclusions

This chapter investigated the topic of decision-making in the maritime domain, emphasizing robust and explainable procedures with computational methods, AD algorithms and RA criteria. It has introduced for this purpose a conceptual modular architecture that encompasses the full scope of a maritime predictive maintenance and fault tolerance control loop. The proposed architecture is a highly generic model that utilizes multiple modalities of inputs, *e.g.,* sensors and streams, for its DAQ module, a data analysis routine with a novel Failure Prediction Block that extracts semantic information from data and performs decision-making tasks and a Failure Prevention block that can aggregate the previous information and raise alarms with explainable outcomes and actuation/healing strategies as well. A literature review was performed to identify alternatives for the implementation of the different modules comprising the closed-loop model, as well as open issues and relevant research challenges. Overall, this highlights that trustworthy maritime decision-making requires the integration of forecasting, AD, RA, and explainability within a closed-loop failure prevention process. In this context, explainability is not an isolated component but a key requirement for making AI-assisted recommendations understandable and actionable for operators and engineers. As these technologies continue to evolve, they create new opportunities to optimize maritime operations, reduce costs, and enhance equipment reliability. DTs can further support this direction by improving flexibility and efficiency in maintenance planning and execution.

**Acknowledgment**
This work has received funding by the European Union's Horizon Europe research and innovation programme under grant agreement No101202933 (D-NAVIO). Views and opinions expressed are however those of the authors only and do not necessarily reflect those of the European Union or the European Climate, Infrastructure, and Environment Executive Agency (CINEA). Neither the European Union nor the granting authority can be held responsible for them.